\documentclass[10pt,twocolumn,letterpaper]{article}

\usepackage{cvpr}            
\usepackage{graphicx}
\usepackage{amsmath}
\usepackage{amssymb}
\usepackage{booktabs}
\usepackage{bm}
\usepackage{multirow}
\usepackage{makecell}

\usepackage[pagebackref,breaklinks,colorlinks]{hyperref}
\usepackage[capitalize]{cleveref}
\crefname{section}{Sec.}{Secs.}
\Crefname{section}{Section}{Sections}
\Crefname{table}{Table}{Tables}
\crefname{table}{Tab.}{Tabs.}

\begin{document}

\title{Deep Image Harmonization with Globally Guided Feature Transformation and Relation Distillation}

\author{Li Niu$^{1}$\thanks{Corresponding author.}~, Linfeng Tan$^{1}$, Xinhao Tao$^{1}$, Junyan Cao$^{1}$,  Fengjun Guo$^{2}$, Teng Long$^{2}$, Liqing Zhang$^{1}$ \\
$^{1}$ Department of Computer Science and Engineering, MoE Key Lab of Artificial Intelligence, \\
Shanghai Jiao Tong University\\
$^{2}$ INTSIG \\
{\tt \scriptsize \{ustcnewly,tanlinfeng,taoxinhao,Joy\_C1\}@sjtu.edu.cn, \{fengjun\_guo,mike\_long\}@intsig.net,  zhang-lq@cs.sjtu.edu.cn}
}

\maketitle

\begin{abstract}
Given a composite image, image harmonization aims to adjust the foreground illumination to be consistent with background. Previous methods have explored transforming foreground features to achieve competitive performance. In this work, we show that using global information to guide foreground feature transformation could achieve significant improvement. Besides, we propose to transfer the foreground-background relation from real images to composite images, which can provide intermediate supervision for the transformed encoder features. Additionally, considering the drawbacks of existing harmonization datasets, we also contribute a ccHarmony dataset which simulates the natural illumination variation. Extensive experiments on iHarmony4  and our contributed dataset demonstrate the superiority of our method. Our ccHarmony dataset is released at
\href{https://github.com/bcmi/Image-Harmonization-Dataset-ccHarmony}{https://github.com/bcmi/Image-Harmonization-Dataset-ccHarmony}.

\end{abstract}


\section{Introduction} \label{sec:intro}

Image composition \cite{niu2021making,chen2019toward}  is an essential editing operation to combine regions from different images to produce composite images, which has a variety of vision applications like advertisement propaganda and digital entertainment \cite{ren2022semantic,liang2021spatial}. Nevertheless, when pasting the foreground extracted from one image on another background image, the resultant composite image may have inconsistent illumination statistics. Image harmonization \cite{CongDoveNet2020,ling2021region} aims to adjust the foreground illumination for visual consistency within the composite image. In recent years, deep learning based image harmonization methods \cite{CongDoveNet2020,sofiiuk2021foreground,Hao2020bmcv,ling2021region,Jiang_2021_ICCV,guo2021image} have sprouted out and achieved remarkable progress. 

Existing methods \cite{peng2022frih,zhu2022image,xiaodong2019improving,ren2022semantic,ke2022harmonizer} have developed myriads of techniques to address the mismatch between foreground and background illumination. Among them, some works~\cite{ling2021region,xiaodong2019improving} revealed that specially transforming the foreground features can enhance the performance, since the main goal of image harmonization is adjusting the foreground. For example, \cite{ling2021region,hang2022scs} transfer the feature statistics (\emph{e.g.}, mean, variance) from background to foreground. \cite{xiaodong2019improving,sofiiuk2021foreground} process foreground and background features separately using different channel attentions. These methods only leverage the information within the same feature map to transform the foreground features, lacking the guidance of global information. However,  the global information of the whole composite image is very likely to be beneficial for foreground feature transformation. 

In this work, we extract the bottleneck feature from encoder as the global feature, which is used to guide the transformation of foreground features in each feature map. Regarding the transformation manner, we opt for modulated convolution kernel proposed in \cite{karras2020analyzing}, and other transformation manners are also applicable. In particular, we use global feature to obtain the modulated convolution kernel, which is applied to the foreground features. We name our method as GiftNet (\textbf{g}lobally gu\textbf{i}ded \textbf{f}eature \textbf{t}ransformation). It is worth noting that our method shares similar spirit with CDTNet~\cite{cong2021high}. \textbf{CDTNet uses global feature to predict color transformation, while our GiftNet uses global feature to predict feature transformation.}

In practice, we observe that globally guided feature transformation is effective for decoder features, but ineffective for encoder features. We conjecture that one reason is the lack of intermediate supervision for encoder features. Specifically, most of previous harmonization methods \cite{peng2022frih,zhu2022image,xiaodong2019improving,ren2022semantic,ke2022harmonizer} only use the ground-truth image as the final supervision, without any intermediate supervision. In the auto-encoder based network, the encoder features may not be sufficiently harmonized, which could have negative impact on the harmonization performance. To address the above issue, we provide useful guidance for encoder features, by distilling foreground-background relation from the encoder feature maps of harmonious images to those of composite images. In particular, we design a two-branch network, in which the top branch reconstructs the harmonious real images and the bottom branch harmonizes the inharmonious composite image. We add a distillation loss to pull close the foreground-background relation in encoder feature maps between two branches.

Another contribution of this work is a new harmonization dataset. Training deep harmonization models requires abundant pairs of composite images and harmonious images. 
Due to the high cost of manually harmonizing composite images, the prevalent way to construct harmonization dataset is the inverse approach \cite{CongDoveNet2020,ren2022semantic,xing2022composite}, \emph{i.e.}, adjusting the foreground of real images to create synthetic composite images. However, this manner may not faithfully reflect natural illumination variation, that is, the same foreground object captured under different illumination conditions. To faithfully 
reflect natural illumination variation, we need to capture a group of images for the same scene under varying illumination conditions, as Hday2night subdataset in iHarmony4 \cite{CongDoveNet2020}. However, such data collection process is extremely expensive. \textbf{In this work, we propose a novel way to construct harmonization dataset, aiming to simulate natural illumination variation.} Specifically, we utilize existing datasets \cite{cheng2014illuminant,gehler2008bayesian}, in which each image is associated with its illumination information recorded by color checker (see Figure~\ref{fig:ccHarmony_illustration}). Based on the recorded illumination information, we can transfer each image across different illumination conditions to simulate natural illumination variation. We name our dataset as \textbf{c}olor-\textbf{c}hecker harmonization (ccHarmony) dataset, which offers a new perspective to construct harmonization dataset.

Our contributions can be summarized as follows. 1) We propose globally guided feature transformation to adjust the foreground features, which demonstrates the importance of global guidance for foreground feature transformation; 2) We propose to distill foreground-background relation from harmonious feature map to composite feature map, which provides useful intermediate supervision for image harmonization task. 3) We contribute a new dataset named ccHarmony to approximate natural illumination variation, which offers a new perspective for harmonization dataset construction. 
4) Extensive experiments on the benchmark dataset iHarmony4~\cite{CongDoveNet2020} and our contributed dataset show that our method significantly outperforms the existing methods. 

\section{Related Work}

\subsection{Image Harmonization}
Image harmonization \cite{CongDoveNet2020,cong2021high,PHDNet} aims to harmonize a composite image by adjusting foreground illumination to match background illumination. 
Early image harmonization methods
\cite{song2020illumination,xue2012understanding,multi-scale,lalonde2007using} used traditional color matching algorithms to match the low-level color statistics between foreground and background.
In recent years, lots of supervised deep harmonization methods \cite{tsai2017deep,Jiang_2021_ICCV,xing2022composite,peng2022frih,zhu2022image,valanarasu2022interactive,bao2022deep,cao2021deep} have emerged. For example, \cite{xiaodong2019improving,Hao2020bmcv,sofiiuk2021foreground} proposed diverse attention modules to treat the foreground and background separately, or establish the relation between foreground and background. \cite{CongDoveNet2020,bargain,ling2021region,hang2022scs} directed image harmonization to domain translation or style transfer by treating different illumination conditions as different domains or styles.  \cite{guo2021image,guo2021intrinsic,guo2022transformer} introduced Retinex theory \cite{land1971lightness,land1977retinex} to image harmonization task by decomposing an image into reflectance map and illumination map.  More recently, \cite{cong2021high,ke2022harmonizer,liang2021spatial,xue2022dccf} used deep network to predict color transformation, striking a good balance between efficiency and effectiveness. Different from the above works, we explore global guidance for feature transformation and effective supervision for intermediate features. 

\subsection{Image Harmonization Datasets}
Previous works have explored various ways to construct image harmonization dataset.  1) \cite{CongDoveNet2020} contributes iHarmony4 dataset by adjusting the foregrounds in real images to generate synthetic composite images. However, such color adjustment may not faithfully reflect natural illumination variation. 2) The Hday2night subdataset in \cite{CongDoveNet2020} captures a group of images for the same scene under different illumination conditions, which can reflect natural illumination variation. Nevertheless, such data collection process is extremely expensive. 
3) \cite{Jiang_2021_ICCV} constructed RealHM by manually harmonizing the composite images, which is extremely expensive and not scalable. Moreover, the manually harmonized images are subjective and may not be reliable.
4) Other works \cite{guo2021intrinsic,cao2021deep,bao2022deep} use 3D render to render the same 3D scene or the same 3D foreground object with different illumination conditions, resulting in a group of images. Then, they swap foregrounds within the same group to construct pairs of composite images and ground-truth images.  
However, rendered images have huge domain gap with real images, which may not be helpful when we have adequate real training data \cite{cao2021deep}. In this work, we propose a novel way to approximate natural illumination variation, based on the images with recorded illumination information.

\subsection{Knowledge Distillation}

Knowledge distillation \cite{hinton2015distilling} targets at transferring knowledge from teacher network to student network. Based on the type of knowledge, knowledge distillation can be divided into three groups \cite{gou2021knowledge}: response-based \cite{hinton2015distilling}, feature-based \cite{Li_2017_CVPR}, and relation-based \cite{Yim_2017_CVPR}.
Response-based knowledge distillation usually uses the neural response of the last output layer of the teacher model to supervise the student model \cite{hinton2015distilling,zhang2019fast}.
Feature-based knowledge distillation \cite{romero2014fitnets,wang2020exclusivity,wang2021distilling} targets at matching the feature activations between teacher network and  student network. Relation-based knowledge distillation \cite{Yim_2017_CVPR,passalis2020heterogeneous} pays more attention to the relation between different model layers or data samples. Our work belongs to relation distillation. Precisely, we distill foreground-background relation from reconstruction network to harmonization network.

\section{Our Method}

\begin{figure*}[t]
\begin{center}
\includegraphics[width=0.95\linewidth]{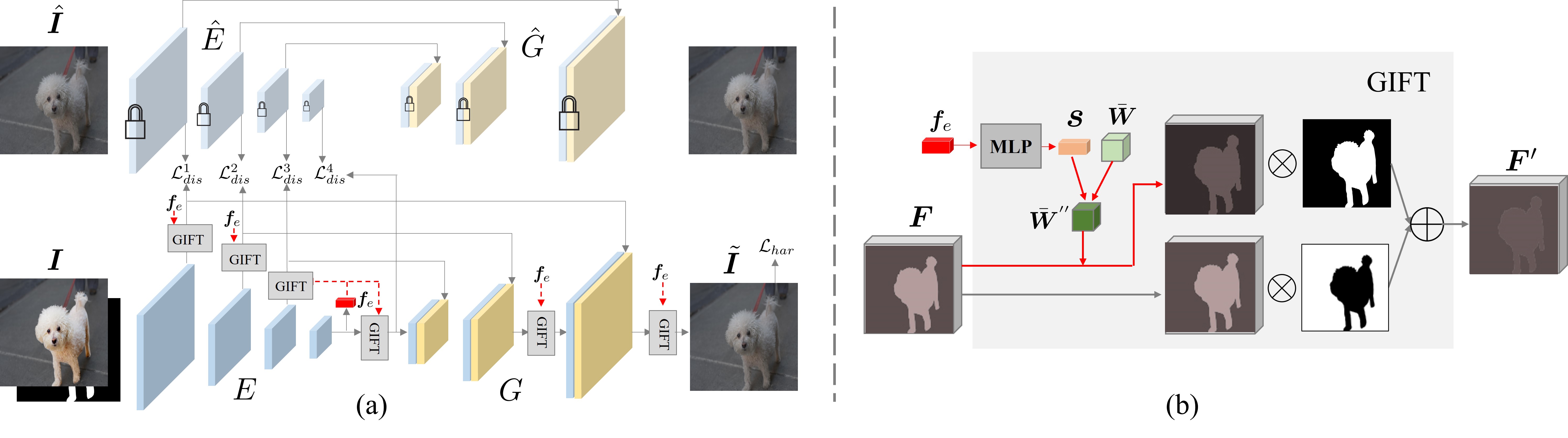}
\end{center}
\caption{(a) Our network which consists of a reconstruction branch ($\hat{E}, \hat{G}$) and a harmonization branch ($E, G$). The  reconstruction branch is pretrained and fixed. In the harmonization branch, we use global feature $\bm{f}_e$ to guide feature transformation using our GIFT module. We also distill foreground-background relation from reconstruction branch to harmonization branch. (b) The detailed architecture of our GIFT module. We use global feature $\bm{f}_e$ to predict scale vector $\bm{s}$ and obtain the modulated convolution weights $\bar{\bm{W}}''$, which are used to adjust the foreground feature map while the background feature map remains unchanged.   }
\label{fig:network}
\end{figure*}

We denote the composite image as $\bm{I}$, which is composed of foreground $\bm{I}^f$ and background $\bm{I}^b$. The ground-truth real image is $\hat{\bm{I}}$. The goal of image harmonization task is adjusting the composite foreground $\bm{I}^f$ and produce the harmonized image $\tilde{\bm{I}}$, which should be close to $\hat{\bm{I}}$.
Our method is built upon the UNet-like network used in \cite{sofiiuk2021foreground}, which consists of an encoder $E$ with four encoder blocks, a decoder $G$ with three decoder blocks, and skip connections from the first three encoder blocks to the corresponding decoder blocks. 
We insert our designed \textbf{g}lobally gu\textbf{i}ded \textbf{f}eature \textbf{t}ransformation (GIFT) module into the network, which will be detailed in Section \ref{sec:GIFT}. Then, we introduce another reconstruction branch to distill the relation knowledge to the harmonization branch, which will be detailed in Section \ref{sec:relation_distillation}.

\subsection{Globally Guided Feature Transformation} \label{sec:GIFT}

We design a \textbf{g}lobally gu\textbf{i}ded \textbf{f}eature \textbf{t}ransformation (GIFT) module, which transforms the foreground feature map using global guidance. We use $\bm{F}_{e,l}$ (\emph{resp.}, $\bm{F}_{d,l}$) to denote the output feature map from the $l$-th encoder (\emph{resp.}, decoder) block. There are four encoder blocks and thus $\bm{F}_{e,4}$ is the bottleneck feature map. 
We perform global average pooling over $\bm{F}_{e,4}$ and get the global feature vector $\bm{f}_e$, which encodes the global information beneficial for image harmonization task. We use $\bm{f}_e$ to guide the feature transformation for encoder/decoder feature maps.  We adopt the feature transformation technique proposed in \cite{karras2020analyzing}, which applies modulated convolution weights to the given feature map. Other feature transformation techniques are also applicable. In our task, we use global feature $\bm{f}_e$ to guide the modulation of convolution weights, which provides global guidance for feature transformation. Besides, we only transform the foreground feature map, since image harmonization aims to adjust the foreground to be compatible with the background.    

Without loss of generality, we take one encoder/decoder feature map as an example. We have learnable base convolution weights $\bar{\bm{W}}$, in which each entry $\bar{W}(m,n,p)$ is the weight for the $m$-th input channel, $n$-th output channel, and the $p$-th spatial location. Then, we pass $\bm{f}_e$  through an Multi-Layer Perceptron (MLP) to predict the scale vector $\bm{s}$ corresponding to input channels. Our used MLP contains four fully-connected layers, in which the first three are shared by all feature maps and the last one is specific for each feature map. The predicted scale vector is used to modulate the base convolution weights: 
\begin{eqnarray}
\bar{W}'(m,n,p) = \bar{W}(m,n,p)\cdot s(m), 
\end{eqnarray}
where $s(m)$ is the $m$-th entry in $\bm{s}$ (the scale for the $m$-th input channel), and $\bar{W}'(m,n,p)$ is the modulated weight. Next, we normalize the modulated weights following \cite{karras2020analyzing}:
\begin{eqnarray}
\bar{W}''(m,n,p) = \frac{\bar{W}'(m,n,p)}{\sqrt{{\sum_{m,p} \bar{W}'(m,n,p)}^2+\epsilon}},
\end{eqnarray}
in which $\epsilon$ is a small constant to avoid numerical error. For more details of convolution weights modulation, please refer to \cite{karras2020analyzing}. Assuming that the feature map to be transformed is $\bm{F}$, which consists of the foreground feature map $\bm{F}^f$ and background feature map $\bm{F}^b$. As illustrated in Figure~\ref{fig:network}(b), 
we apply the modulated convolution weights $\bar{\bm{W}}''$ to the foreground feature map $\bm{F}^f$ and produce 
the transformed foreground feature map  ${\bm{F}^{f}}'$. Then, we combine ${\bm{F}^{f}}'$ and $\bm{F}^b$ to obtain the transformed feature map  $\bm{F}'$.

We try applying our designed GIFT module to the output feature map from each encoder block or decoder block. We observe that the last two decoder blocks can obviously manifest the advantage of our designed GIFT, while the encoder blocks do not show clear advantage (see Section~\ref{sec:ablation_studies}). We conjecture that the whole network is only supervised by the final ground-truth image and there is no intermediate supervision for the encoder feature maps. Lack of intermediate supervision might hinder the potential of our designed GIFT module. To provide intermediate supervision for the encoder features, we borrow the idea from knowledge distillation, which will be introduced next.

\subsection{Relation Distillation} \label{sec:relation_distillation}

The difficulty of supervising encoder feature maps lies in the absence of ground-truth harmonized encoder feature maps. One intuitive thought is that the harmonized encoder feature maps should be close to the encoder feature maps of ground-truth real images. Therefore, besides the harmonization branch, we introduce another reconstruction branch which reconstructs the ground-truth real image $\hat{\bm{I}}$. The reconstruction branch shares similar UNet network structure (encoder $\hat{E}$ and decoder $\hat{G}$) with the harmonization branch. We denote the $l$-th encoder feature map in the reconstruction branch as $\hat{\bm{F}}_{e,l}$. Note that we perform distillation on the encoder feature maps $\bm{F}'_{e,l}$ transformed by our GIFT module. 

One naive approach is feature distillation, which enforces the encoder feature maps between two branches to be close, that is, $\hat{\bm{F}}_{e,l}$ is close to $\bm{F}'_{e,l}$ for $l=1,\ldots,4$. However, the performance using feature distillation is unsatisfactory (see Section \ref{sec:ablation_studies}). One possible reason is that the encoder feature maps contain lots of redundant information and distilling the entire feature maps increases the difficulty of network training. Therefore, we turn to only distill the key information: foreground-background relation. Distilling foreground-background relation can ensure that the foreground feature map is compatible with the background feature map in $\bm{F}'_{e,l}$, as in $\hat{\bm{F}}_{e,l}$. 

There could be many possible ways to characterize foreground-background relation. To avoid the heavy computational cost of pixel-to-pixel relation, we first calculate the averaged foreground feature and then calculate the similarity between it and pixel-wise features. By taking $\bm{F}'_{e,l}$ as an example, we calculate the averaged foreground feature $\bm{f}'_{e,l}$ by performing average pooling within the foreground region. Then, we calculate the similarity between $\bm{f}'_{e,l}$ and the $i$-th pixel-wise feature in $\bm{F}'_{e,l}$ as follows,
\begin{eqnarray}
R'_l[i] = \frac{\exp(-\gamma \|\bm{f}'_{e,l}- \bm{F}'_{e,l}[i]\|^2)}{\sum_{j} \exp(-\gamma \|\bm{f}'_{e,l} - \bm{F}'_{e,l}[j]\|^2)},
\end{eqnarray}
in which $\gamma$ is a hyper-parameter. 
The similarities $R'_l[i]$ of all pixels form a similarity map $\bm{R}'_l$, in which the values in the background region represent the relation between foreground and background. The background regions with similar appearance to the foreground should have higher values. 
The similarity map $\hat{\bm{R}}_l$ for $\hat{\bm{F}}_{e,l}$ could be calculated in a similar way. Then, we distill foreground-background relation from $\hat{\bm{F}}_{e,l}$ to $\bm{F}'_{e,l}$ using the distillation loss
$\mathcal{L}^l_{dis} = \|\bm{R}'_l-\hat{\bm{R}}_l\|^2$,
which pushes the foreground-background relation in the harmonized feature map towards that in the real feature map, so that foreground is compatible with the background in the harmonized feature map. 

During training, we first train and fix the reconstruction branch. Then, we train the harmonization branch. In addition to the distillation loss, we enforce the harmonized image $\tilde{\bm{I}}$ to be close to the ground-truth image $\hat{\bm{I}}$ by $\mathcal{L}_{har} = \|\tilde{\bm{I}}-\hat{\bm{I}}\|_1$. 

After summing up the distillation losses for four encoder blocks, the total loss for the harmonization branch is
\begin{eqnarray} \label{eqn:loss_total}
\mathcal{L}_{all} = \mathcal{L}_{har} +\lambda \sum_{l=1}^4 \mathcal{L}^l_{dis},
\end{eqnarray}
in which $\lambda$ is a hyper-parameter. 

\begin{figure*}[t]
\begin{center}
\includegraphics[width=0.8\linewidth]{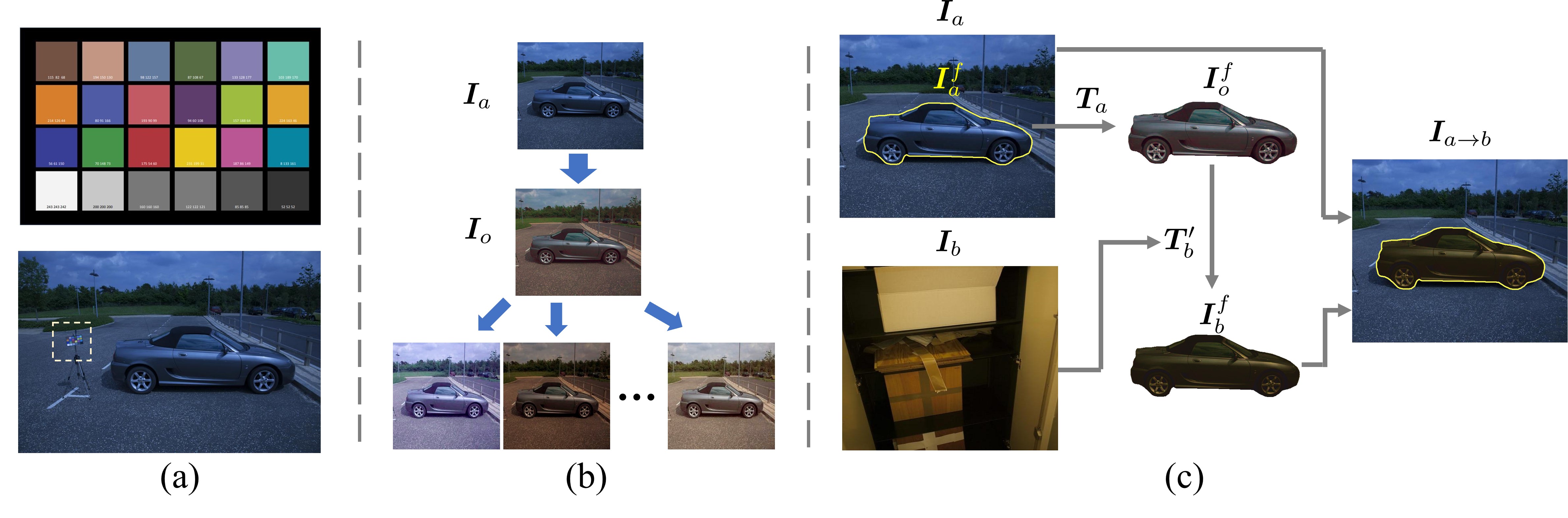}
\end{center}
\vspace{-10pt}
\caption{(a) A color checker in standard illumination condition and an image captured with color checker (see the yellow bounding box). (b) We first convert $\bm{I}_a$ to $\bm{I}_o$ in standard illumination condition, and then convert $\bm{I}_o$ to other illumination conditions.  (c) The construction process of our dataset.
Given a real image $\bm{I}_a$, we convert its foreground $\bm{I}_a^f$ to $\bm{I}_o^f$ in standard illumination condition using polynomial matching matrix $\bm{T}_a$. Then, we convert $\bm{I}_o^f$ to $\bm{I}_b^f$ in the illumination condition of reference image $\bm{I}_b$ using the inverse polynomial matching matrix $\bm{T}'_b$. Finally, $\bm{I}_b^f$ is combined with the background of $\bm{I}_a$ to produce a synthetic composite image $\bm{I}_{a\rightarrow b}$.}
\label{fig:ccHarmony_illustration}
\end{figure*}

\section{Our ccHarmony Dataset}

In this work, we explore a novel transitive way to construct image harmonization dataset, aiming to simulate the natural illumination variation. Specifically, based on the existing datasets with recorded illumination information, we first convert the foreground in a real image to the standard illumination condition, and then convert it to another illumination condition, which is combined with the original background to produce a synthetic composite image. 

We build our dataset upon two publicly datasets with recorded illumination information: NUS dataset \cite{cheng2014illuminant} and Gehler dataset \cite{gehler2008bayesian}, which were originally constructed for the research on color constancy. In these two datasets, each image is captured with a color checker placed in the scene that provides ground-truth reference for illumination estimation. Given a 24-patch Macbeth color checker, we have the original colors of 24 patches in standard illumination condition (see Figure \ref{fig:ccHarmony_illustration}(a)), which is referred to as standard patch color. Given an image with color checker, we can extract the colors of 24 patches, which is referred to as image patch color.

Inspired by previous works \cite{zhou2018color,luo2021estimating,afifi2019color}, we use polynomial matching to characterize the color transformation between standard patch colors and image patch colors. Formally, we denote the standard patch colors of 24 patches as $\mathcal{C}_o=\{\bm{c}^o_1,\bm{c}^o_2,\ldots,\bm{c}^o_{24}\}$ and image patch colors of 24 patches in image $\bm{I}_a$ as  $\mathcal{C}_a=\{\bm{c}^a_1,\bm{c}^a_2,\ldots,\bm{c}^a_{24}\}$. We can learn a polynomial matching matrix $\bm{T}_a$ which converts $\mathcal{C}_a$ to $\mathcal{C}_o$.
$\bm{T}_a$ could convert $\bm{I}_a$ to $\bm{I}_o$ in standard illumination condition.
We can also learn an inverse polynomial matching matrix $\bm{T}'_a$ which converts $\mathcal{C}_o$ to $\mathcal{C}_a$.
Provided with inverse polynomial matching matrices from other images, we can convert $\bm{I}_o$ to the illumination conditions of other images. As shown in Figure~\ref{fig:ccHarmony_illustration}(b), with  standard illumination condition being the transitional state, we can transfer across different illumination conditions to simulate natural illumination variation.

Next, we briefly describe our way of creating synthetic composite images. As depicted in Figure~\ref{fig:ccHarmony_illustration}(c), given an image $\bm{I}_a$ with foreground mask, we convert its foreground $\bm{I}_a^f$ to the one $\bm{I}_o^f$ in standard illumination condition using polynomial matching matrix $\bm{T}_a$. Next, by using the inverse polynomial matching matrix $\bm{T}'_b$ of another reference image $\bm{I}_b$, we convert $\bm{I}_o^f$ to $\bm{I}_b^f$ in the illumination condition of $\bm{I}_b$. Finally, we replace the foreground  $\bm{I}_a^f$ in $\bm{I}_a$ with its counterpart $\bm{I}_b^f$, yielding a synthetic composite image $\bm{I}_{a\rightarrow b}$. In this manner, we can acquire adequate pairs of synthetic composite images and real images like $\{\bm{I}_{a\rightarrow b}, \bm{I}_a\}$. Because our dataset is constructed based on images with color checker (cc), we name our dataset as ccHarmony. The details of our dataset construction are left to supplementary due to space limitation. 

\section{Experiments}

\subsection{Datasets and Implementation Details} 

\begin{table*}[t]
\small
\centering
\setlength{\tabcolsep}{1mm}{%
\begin{tabular}{c|c|c|c|c|c|c|c|c|c|c|c|c|c|c|c}
\hline
 \multirow{2}{*}{Method} & \multicolumn{3}{c|}{All}  & \multicolumn{3}{c|}{HCOCO} & \multicolumn{3}{c|}{HFlickr} & \multicolumn{3}{c|}{HAdobe5k} & \multicolumn{3}{c}{Hday2night} \\  
\cline{2-16}
 ~ & MSE & fMSE & PSNR & MSE & fMSE & PSNR & MSE & fMSE & PSNR & MSE & fMSE & PSNR & MSE & fMSE & PSNR
 \\ \hline
  DoveNet~\cite{CongDoveNet2020} & 52.36 & 549.96 & 34.75 & 36.72 & 554.55 & 35.83 & 133.14 & 832.64 & 30.21  & 52.32 & 383.91 & 34.34 & 54.05 & 1075.42 & 35.18 \\ 
 RainNet~\cite{ling2021region} & 40.29 & 469.60 & 36.12 & 31.12 & 535.40 & 37.08 & 117.59 & 751.12 & 31.64 & 42.85 & 320.43 & 36.22 & 47.24 & 852.12 & 34.83 \\
 IIH~\cite{guo2021intrinsic} & 38.71 & 400.29 & 35.90 & 24.92 & 416.38 & 37.16 & 105.13 & 716.60 & 31.34 & 43.02 & 284.21 & 35.20 & 55.53 & 797.04 & 35.96 \\
 IHT~\cite{guo2021image} & 27.89 & 295.56 & 37.94 & 14.98 & 274.67 & 39.22 & 67.88 & 471.04 & 33.55 & 36.83 & 242.57 & 37.17 & 49.67 & 736.55 & 36.38 \\
 iSSAM~\cite{sofiiuk2021foreground} & 24.64 & 262.67 & 37.95 & 16.48 & 266.14 & 39.16 & 69.68 & 443.63 & 33.56 & 22.59 & 166.19 & 37.24 & 40.59 & 591.07 & 37.72 \\
 CDTNet~\cite{cong2021high} & 23.75 & 252.05 & 38.23 & 16.25 & 261.29 & 39.15 & 68.61 & 423.03 & 33.55 & 20.62 & 149.88 & 38.24 & 36.72 & 549.47 & \textbf{37.95} \\
 Harmonizer~\cite{ke2022harmonizer} & 24.26 & 280.51 & 37.84 & 17.34 & 298.42 & 38.77 & 64.81 & 434.06 & 33.63 & 21.89& 170.05 & 37.64 & \textbf{33.14} & \textbf{542.07} & 37.56 \\
 DCCF~\cite{xue2022dccf} & 22.05 & 266.49 & 38.50 & 14.87 & 272.09 & 39.52 & 60.41 & 411.53 & 33.94 & 19.90 & 175.82 & 38.27 & 49.32 & 655.43 & 37.88 \\
 \hline
 GiftNet & \textbf{19.46} & \textbf{225.30} & \textbf{38.92} & \textbf{12.70} & \textbf{229.68} & \textbf{39.91} & \textbf{54.33} & \textbf{360.08} & \textbf{34.44} & \textbf{18.35} & \textbf{143.96} & \textbf{38.76} & 38.28 & 566.47 & 37.81\\
 \hline
\end{tabular}
}
\caption{Quantitative comparison on iHarmony4 \cite{CongDoveNet2020} dataset. The best results are denoted in boldface.}
\label{tab:results_iHarmony4}
\end{table*}

We conduct experiments on the benchmark iHarmony4 \cite{CongDoveNet2020} and our constructed ccHarmony dataset.  iHarmony4 \cite{CongDoveNet2020} is composed of four sub-datasets: HCOCO, HFlickr, HAdobe5K,  Hday2night, which contains totally $65,742$ (\emph{resp.}, $7,404$) pairs of composite images and ground-truth real images in the training (\emph{resp.}, test) set. Following previous works \cite{CongDoveNet2020,sofiiuk2021foreground}, we merge the training sets of four subdatasets as the whole training set and evaluate on each subdataset. Our constructed ccHarmony dataset has $3,080$ (\emph{resp.}, $1,180$) pairs of composite images and ground-truth real images in the training (\emph{resp.}, test) set. For the experiments on ccHarmony, we first pretrain the model on iHarmony4 \cite{CongDoveNet2020} and then finetune the model on ccHarmony, in which the input image size is set as $256\times 256$.

For evaluation metrics, following previous works \cite{CongDoveNet2020,sofiiuk2021foreground,guo2021intrinsic,cong2021high}, we adopt PSNR, MSE, fMSE, fSSIM, in which fMSE (\emph{resp.}, fSSIM) means MSE (\emph{resp.}, SSIM) within the foreground region. 
Our model is implemented using Pytorch 1.10.1 and trained using Adam optimizer with learning rate being $1e{-3}$ on ubuntu 20.04 operation system, Intel(R) Xeon(R) Silver 4116 CPU, and RTX 3090 GPUs with 24GB memory. We empirically set $\gamma$ as $0.01$ and $\lambda$ as $0.001$.

\subsection{Comparison with Baselines}
On iHarmony4, we compare with following image harmonization methods: DoveNet~\cite{sofiiuk2021foreground}, RainNet~\cite{ling2021region}, IIH~\cite{guo2021intrinsic}, IHT~\cite{guo2021image}, iSSAM~\cite{sofiiuk2021foreground}, CDTNet~\cite{cong2021high}, Harmonizer~\cite{ke2022harmonizer}, DCCF~\cite{xue2022dccf}. In Table \ref{tab:results_iHarmony4}, we report the results on four sub test sets and the whole test set, which are copied from original papers or reproduced with the released models.  For the overall results on the whole test set, our method outperforms the SOTA method by a large margin, \emph{e.g.}, 10.61\% relative improvement over CDTNet on fMSE and 11.75\% relative improvement over DCCF on MSE. Our method achieves the best results on HCOCO, HFlickr, HAdobe5k. Our method does not achieve satisfactory results on Hday2night, probably due to the limited training set (only $311$ images). 

\begin{table}[t]
\centering
\small
\begin{tabular}{c|c|c|c|c}
\hline
 Method  & MSE$\downarrow$ & fMSE$\downarrow$ & PSNR$\uparrow$ & fSSIM $\uparrow$\\ \hline
 DoveNet~\cite{CongDoveNet2020} & 110.84 & 880.94 & 31.61 & 0.8231 \\
 RainNet~\cite{ling2021region} &  58.11 & 519.32 & 34.78 & 0.8665 \\
 IIH~\cite{guo2021intrinsic} & 83.72 & 636.28 & 33.64 & 0.7640 \\
 IHT~\cite{guo2021image} & 55.73 & 514.47 & 35.07 & 0.8203 \\
 iSSAM~\cite{sofiiuk2021foreground} & 28.83 & 264.84 & 36.05 & 0.9096\\
 CDTNet~\cite{cong2021high} & 27.87 & 264.51 & 36.62 & 0.9225 \\
 Harmonizer~\cite{ke2022harmonizer} & 43.31 & 402.09 & 34.68 & 0.8951 \\
 DCCF~\cite{xue2022dccf} & 29.25 & 259.83 & 36.62 & 0.9094 \\
 \hline
 GiftNet & \textbf{24.55} & \textbf{235.20} & \textbf{37.59} & \textbf{0.9322} \\
 \hline
\end{tabular}
\caption{Quantitative comparison on our ccHarmony dataset. The best results are denoted in boldface.}
\label{tab:results_ccHarmony}
\end{table}

On ccHarmony, we also report the results of our method and baselines in Table \ref{tab:results_ccHarmony}. Again, our method outperforms the baselines and achieves significant improvement. 

\begin{figure*}[t]
\begin{center}
\includegraphics[width=0.94\linewidth]{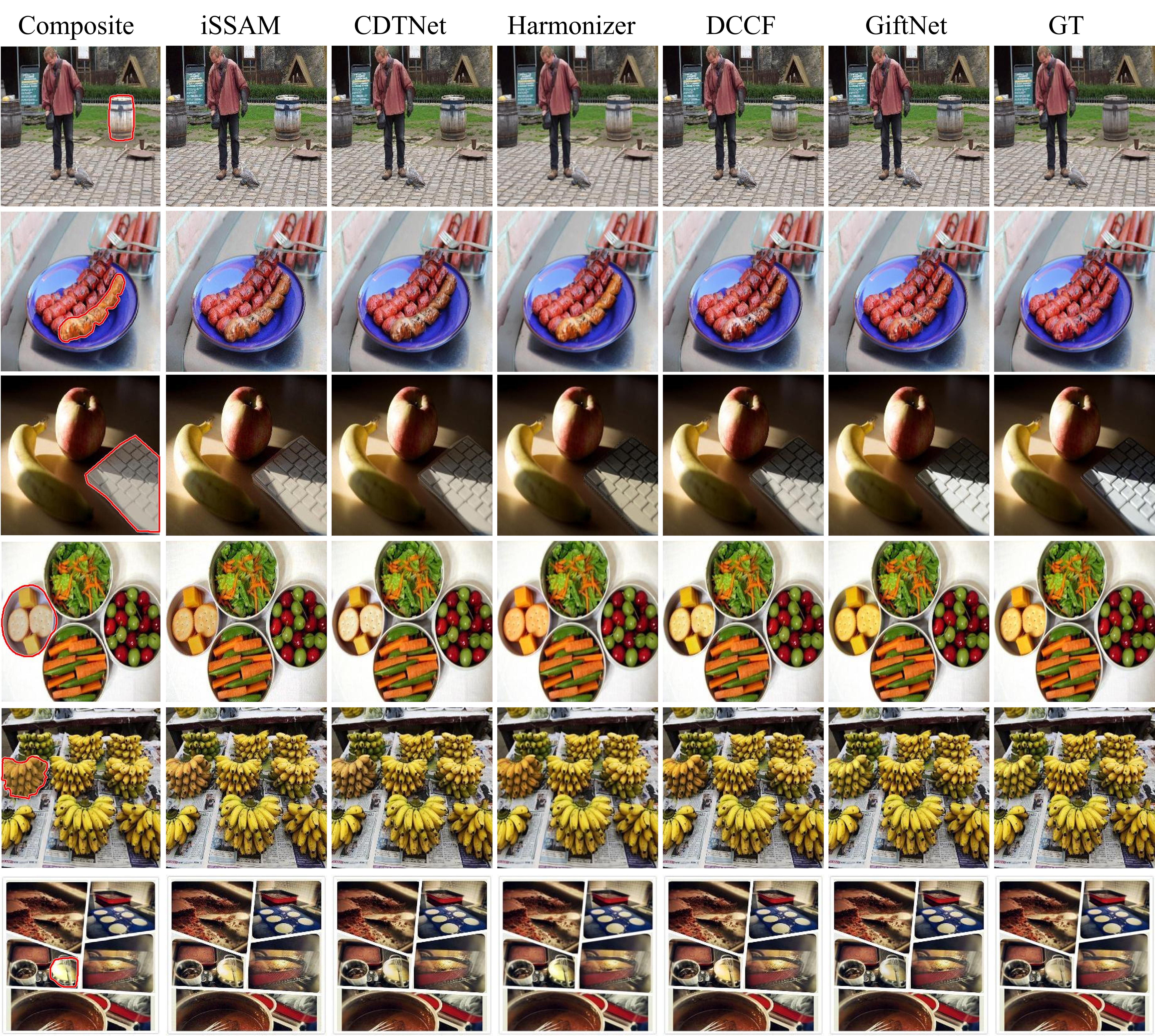}
\end{center}
\caption{From left to right, we show the composite image (foreground outlined in red), the harmonized results of  iSSAM~\cite{sofiiuk2021foreground},  CDTNet~\cite{cong2021high}, Harmonizer~\cite{ke2022harmonizer}, DCCF~\cite{xue2022dccf}, our GiftNet, and the ground-truth on iHarmony4~\cite{CongDoveNet2020} dataset.}
\label{fig:Results_iHarmony4_main}
\end{figure*}

\begin{figure*}[t]
\begin{center}
\includegraphics[width=0.86\linewidth]{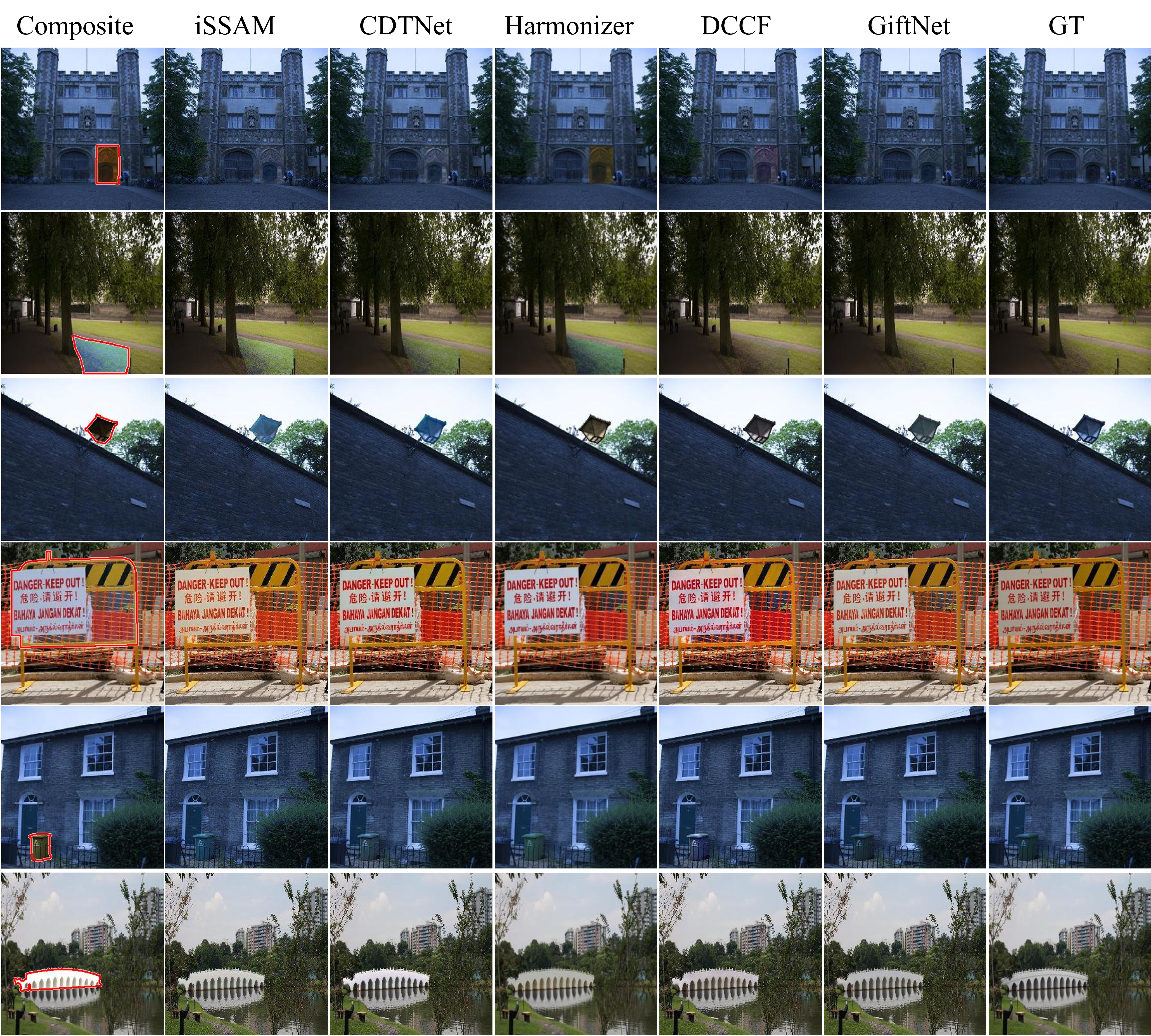}
\end{center}
\caption{From left to right, we show the composite image (foreground outlined in red), the harmonized results of  iSSAM~\cite{sofiiuk2021foreground},  CDTNet~\cite{cong2021high}, Harmonizer~\cite{ke2022harmonizer}, DCCF~\cite{xue2022dccf}, our GiftNet, and the ground-truth on our ccHarmony dataset.}
\label{fig:Results_ccHarmony_main}
\end{figure*}

For the competitive baselines \cite{sofiiuk2021foreground,cong2021high,ke2022harmonizer,xue2022dccf} and our method, we show the harmonized results on iHarmony4 in Figure \ref{fig:Results_iHarmony4_main} and the results on ccHarmony in Figure \ref{fig:Results_ccHarmony_main}. Our method can usually produce visually pleasant results which are closer to the ground-truth real images. More visualization results can be found in the supplementary.

\begin{table}[t]
\centering
\small
\setlength{\tabcolsep}{1.1mm}{
\begin{tabular}{c|c|c|c|c|c|c|c}
\hline
\multirow{2}{*}{Row} &\multicolumn{2}{c|}{GIFT}  & \multicolumn{2}{c|}{Distill} & \multicolumn{3}{c}{Evaluation}  \\  
\cline{2-8}
  & Layer & Op & Layer &  Op & MSE & fMSE & PSNR \\ \hline
  1 & &   & & & 24.64 & 262.67 & 37.95 \\
  \hline
 2 & $D_3$ & * & & & 21.39 & 243.23 & 38.71 \\
 3 & $D_{2,3}$ & * & & & 20.82 & 238.89 & 38.76 \\ 
 4 & $D_{1,2,3}$ & * & & & 20.79 & 237.06 & 38.78 \\
 5 & $D_{2,3},E$ & * & & & 20.50 & 236.22 & 38.77 \\ 
 \hline
 6 & $D_3$ & w/o g & & & 24.33 & 263.22 & 38.18 \\  
 7 & $D_3$ & $D_3\!\rightarrow$g & & & 24.21 & 261.15 & 38.28 \\
 \hline
 8 & $D_{2,3},E$ & * & $E$ & * & \textbf{19.46} & \textbf{225.30} & \textbf{38.92} \\
 9 & $D_{2,3},E$ & * & $D_{2,3},E$ & * & 21.31 & 242.58 & 38.73 \\ 
 10 & $D_{2,3},E$ & * & $E$ & FD & 21.58 & 240.18 & 38.67 \\ 
 \hline
\end{tabular}
}
\caption{Ablation studies on iHarmony4 \cite{CongDoveNet2020} dataset. * means the default operation. $D_k$ means the $k$-th decoder block. $E$ means all encoder blocks. ``FD" is short for feature distillation. ``w/o g" means without guidance. ``$D_3\!\rightarrow$g" means using $D_3$ feature map as guidance. The best results are denoted in boldface.}
\label{tab:ablation_studies}
\end{table}

\subsection{Ablation Studies} \label{sec:ablation_studies}
We conduct ablation studies and report the results in Table~\ref{tab:ablation_studies}. 
After removing GIFT module and relation distillation, our network reduces to \cite{sofiiuk2021foreground}, so we include the results of \cite{sofiiuk2021foreground} in row 1 as the performance of basic model.

\noindent\textbf{GIFT Module: } We first investigate the effectiveness of our GIFT module. Recall that we have four encoder blocks and three decoder blocks. We first append GIFT module to each encoder block or decoder block. We observe that appending GIFT module to the last decoder block achieves the largest improvement. For brevity, we only report the performance of appending to the last decoder block ($D_{3}$ in row 2) in Table \ref{tab:ablation_studies}. 
Based on row 2, we further append GIFT module to the penultimate decoder block ($D_{2,3}$ in row 3) and the performance gain is relatively small. Then, we further append GIFT module to the first decoder block ($D_{1,2,3}$ in row 4) or the encoder blocks ($E$ in row 5), but the performance gains are very marginal. 

Based on row 2, we compare with other types of guidance information for modulating convolution weights. The first type is no guidance information, that is, we use the same learnable input channel scales for all images. The performance is only comparable with row 1, which shows that simply adding more convolution layers does not work. 
The second type is using the feature map to be transformed to predict the input channel scales. In the case of $D_3$, we perform average pooling over the last decoder feature map and pass the averaged feature vector through an MLP to predict the input channel scales. Again, there is no obvious improvement over row 1, demonstrating the necessity of using global information to guide feature transformation. 

\noindent\textbf{Relation Distillation: } Based on row 5, we add relation distillation to encoder feature maps and observe slight performance improvement (row 8 \emph{v.s.} row 5). We also try adding relation distillation to decoder feature maps, but the performance becomes worse (row 9 \emph{v.s.} row 8). One possible reason is that the supervision from the final ground-truth image is sufficient to learn effective decoder features, whereas the relation distillation may disturb the learning of decoder features. Therefore, we only employ relation distillation on encoder feature maps. We also compare relation distillation with feature distillation. In particular, we replace relation distillation loss with feature distillation loss. The obtained performance in row 10 is not satisfactory, which demonstrates the superiority of relation distillation. As explained in Section~\ref{sec:intro}, the whole feature maps contain much redundant and noisy information, which may enlarge the training difficulty of network. In contrast, relation distillation solely distills the key information for image harmonization task. 

\begin{table*}
\centering
\begin{tabular}{c|c|c|c|c|c|c}
\hline Method & Composite &iSSAM~\cite{sofiiuk2021foreground} & CDTNet~\cite{cong2021high} & Harmonizer~\cite{ke2022harmonizer} &  DCCF~\cite{xue2022dccf} & GiftNet \\\hline
B-T score & -1.075 & 0.0755 & 0.212 & 0.117 & 0.235 & 0.436\\
\hline
\end{tabular}
\caption{B-T scores of different methods on $100$ real composite images \cite{cong2021high}. }
\label{tab:BT_score}
\end{table*}
\subsection{Real Composite Images}

We additionally compare different methods on $100$ real composite images provided by \cite{cong2021high}. Since there are no ground-truth for these real composite images, we conduct user study to compare with competitive baselines \cite{sofiiuk2021foreground,cong2021high,ke2022harmonizer,xue2022dccf}.
Following \cite{cong2021high}, given each composite image and its $5$ harmonized results from different methods ($4$ baselines and our method), we can create image pairs by randomly choosing $2$ images from $6$ images ($5$ harmonized results and one composite image). Thus, we can construct $1500$ image pairs based on $100$ real composite images.
We ask $50$ users to watch an image pair each time and choose the more harmonious image, leading to $75,000$ pairwise results. Then, we use the Bradley-Terry (B-T) model \cite{bradley1952rank,lai2016comparative} to calculate the global ranking of all methods, as reported in Table \ref{tab:BT_score}. Our method achieves the highest B-T score, which demonstrates the effectiveness of globally guided feature transformation and relation distillation. The visualization results on real composite images are left to the supplementary due to space limitation.

\section{Conclusion}

In this work, we have proposed globally guided feature transformation and relation distillation for image harmonization. 
We have also contributed a new dataset named ccHarmony, which provides a new perspective for harmonization dataset construction. 
Comprehensive experiments have demonstrated the superiority of our method. 

\section*{Acknowledgement} The work was supported by the National Natural Science Foundation of China (Grant No. 62076162), the Shanghai Municipal Science and Technology Major/Key Project, China (Grant No. 2021SHZDZX0102, Grant No. 20511100300). 

{\small
\bibliographystyle{ieee_fullname}
\bibliography{camera_main.bbl}
}

\end{document}


\title{Supplementary for Deep Image Harmonization with Globally Guided Feature Transformation and Relation Distillation}

\author{Li Niu$^{1}$\thanks{Corresponding author.}~, Linfeng Tan$^{1}$, Xinhao Tao$^{1}$, Junyan Cao$^{1}$,  Fengjun Guo$^{2}$, Teng Long$^{2}$, Liqing Zhang$^{1}$ \\
$^{1}$ Department of Computer Science and Engineering, MoE Key Lab of Artificial Intelligence, \\
Shanghai Jiao Tong University\\
$^{2}$ INTSIG \\
{\tt \scriptsize \{ustcnewly,tanlinfeng,taoxinhao,Joy\_C1\}@sjtu.edu.cn, \{fengjun\_guo,mike\_long\}@intsig.net,  zhang-lq@cs.sjtu.edu.cn}
}

\maketitle

In this document, we provide additional materials to support our main paper. In Section~\ref{sec:ccHarmony_construction}, we provide more details of constructing ccHarmony dataset.  In Section~\ref{sec:cmp_with_baseline}, we show more visual comparison results with baselines. In Section~\ref{sec:real_composite}, we evaluate different methods on real composite images. In Section~\ref{sec:ablation_studies}, we provide visualization results of ablation studies. In Section~\ref{sec:failure_cases}, we discuss the limitation of our method.

\section{More Details of ccHarmony Dataset} \label{sec:ccHarmony_construction}
As introduced in the main paper, the existing harmonization datasets \cite{CongDoveNet2020,ren2022semantic,xing2022composite,Jiang_2021_ICCV} may not 
faithfully reflect natural illumination variation.
The Hday2night subdataset in iHarmony4 \cite{CongDoveNet2020} captures a group of images for the same scene under different illumination conditions, which can reflect natural illumination variation. Nevertheless, such data collection is extremely expensive. Therefore, we explore a novel way to construct harmonization dataset ccHarmony to approximate natural illumination variation.
When constructing our ccHarmony dataset, we collect real images with color checker, segment proper foregrounds, and perform color transfer for the foregrounds, yielding synthetic composite images. Next, we will introduce the above three steps: real image selection in Section~\ref{sec:real_image_selection}, foreground segmentation in Section~\ref{sec:foreground_segmentation}, and foreground color transfer in Section~\ref{sec:foreground_color_transfer}. 

\subsection{Real Image Selection} \label{sec:real_image_selection}

\begin{figure}[t]
\begin{center}
\includegraphics[width=0.95\linewidth]{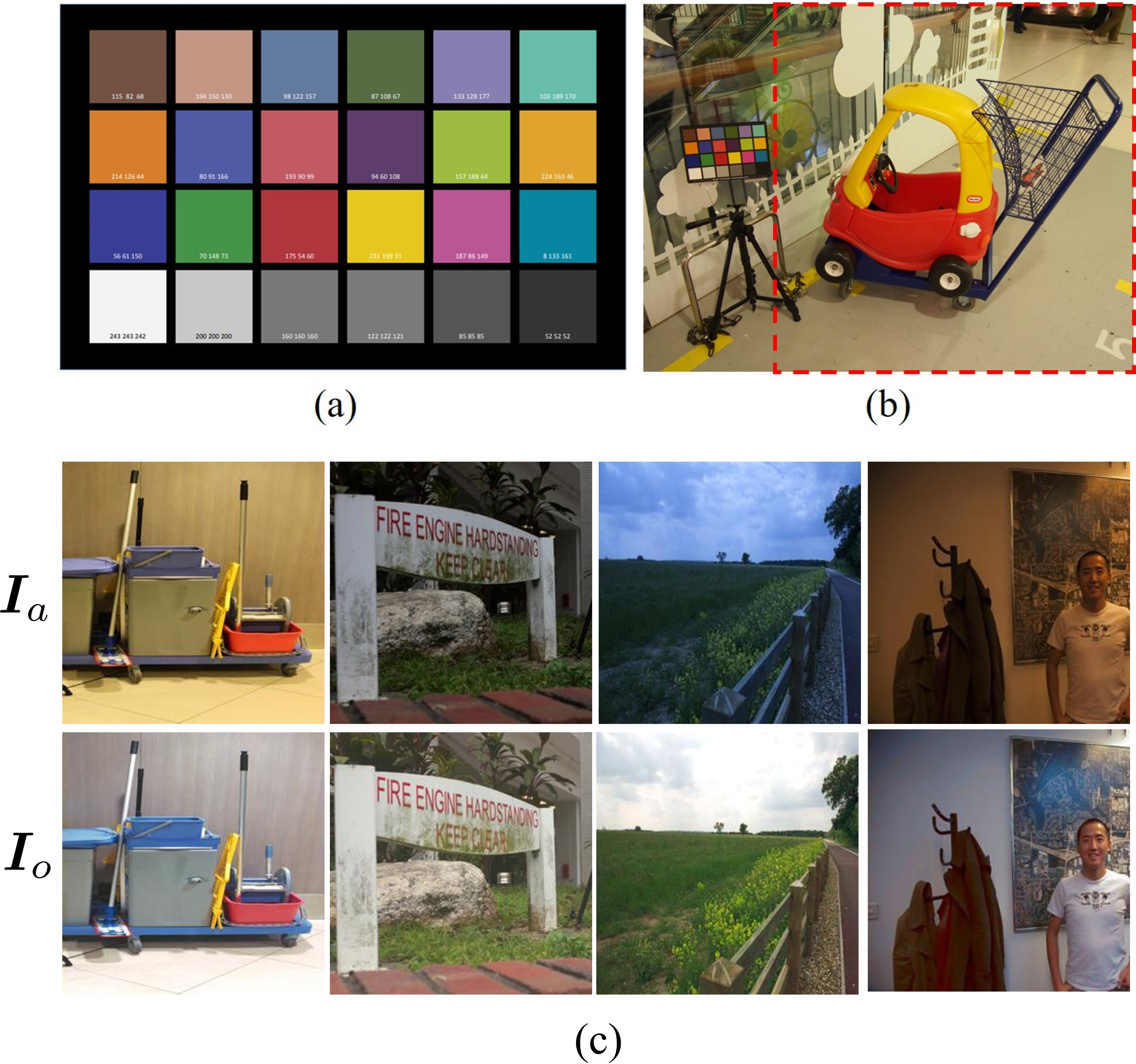}
\end{center}
\caption{(a) Standard patch colors of a 24-patch Macbeth color checker. (b) An image captured with a color checker placed in the scene. The red dashed box indicates the cropped image without color checker, which is used in our dataset. (c) Examples of real images $\bm{I}_a$ and their counterparts $\bm{I}_o$  in standard illumination condition.}
\label{fig:ccHarmony_standard_illumination}
\end{figure}

We first collect images with color checker (see Figure~\ref{fig:ccHarmony_standard_illumination}(a)) from NUS
dataset \cite{cheng2014illuminant} and Gehler dataset \cite{gehler2008bayesian}, in which each image is captured with a color checker placed in the scene to record illumination information. Then, we perform the following filtering steps.
1) We notice that these two datasets contain images capturing the same scene with similar camera viewpoints, so we perform near-duplicate removal to remove the images with duplicated content. 2) We observe that in some images, the color checker cannot represent the global illumination information of the whole image, for example, the color checker is placed in the shadow area. Therefore, we remove those images with misleading color checker. 3) Another issue is that the color checker should not be included in the final image harmonization dataset, because the color check may provide shortcut for the harmonization network. Therefore, we discard the images in which the color checker is placed near the image center, and crop the remaining images to obtain the possibly largest region without color checker (see Figure~\ref{fig:ccHarmony_standard_illumination}(b)).

\begin{figure*}[t]
\begin{center}
\includegraphics[width=0.8\linewidth]{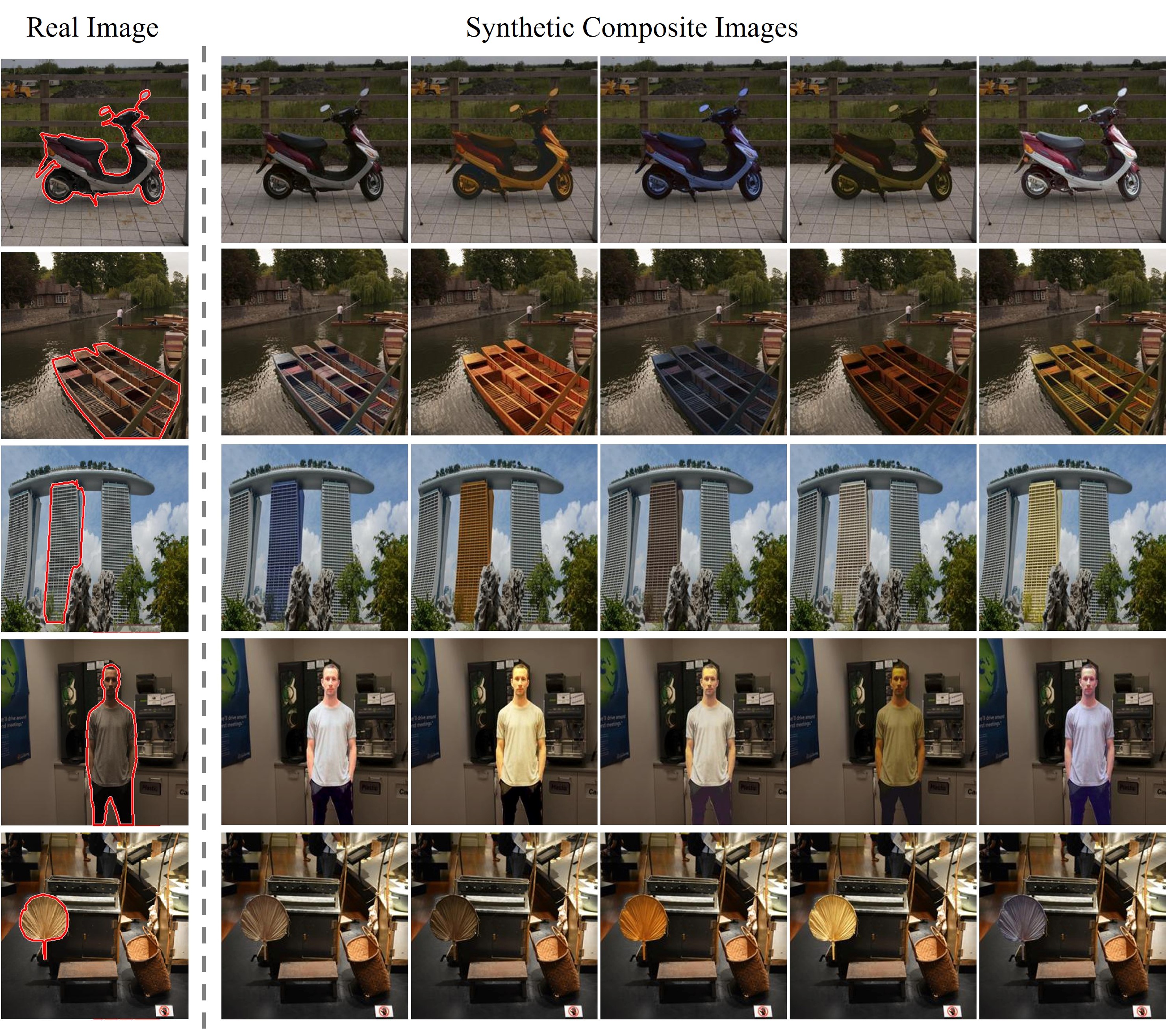}
\end{center}
\caption{We show the real image (foreground outlined in red) in the leftmost column and five example synthetic composite images in the right columns.}
\label{fig:ccHarmony_example_composite}
\end{figure*}

After the above filtering steps, we have $350$ real images. Although the number of real images is limited by the existing datasets \cite{cheng2014illuminant,gehler2008bayesian}, we argue that collecting images with color checker is more scalable than the way of constructing Hday2night~\cite{CongDoveNet2020}, that is, using a fixed camera to capture the same scene over a long time span. Moreover, the main focus of this paper is exploring a novel way to construct harmonization dataset instead of building a large-scale dataset. This work has proved the feasibility of constructing harmonization dataset in this way, and 
ccHarmony could be easily extended by capturing more images with color checkers in the future.
 
\subsection{Foreground Segmentation}\label{sec:foreground_segmentation}
 
For each real image, we manually segment one or two foregrounds. 
When selecting foregrounds, we ensure that the color checker can roughly represent the illumination information of the foreground, so that it is meaningful to apply the polynomial matching matrix calculated based on the color checker to the foreground. In total, we segment $426$ foregrounds in $350$ real images, in which the foregrounds cover a wide range of categories like human, tree, building, furniture, staple goods, and so on (see Figure \ref{fig:ccHarmony_example_composite}). 

\begin{figure*}[t]
\begin{center}
\includegraphics[width=0.98\linewidth]{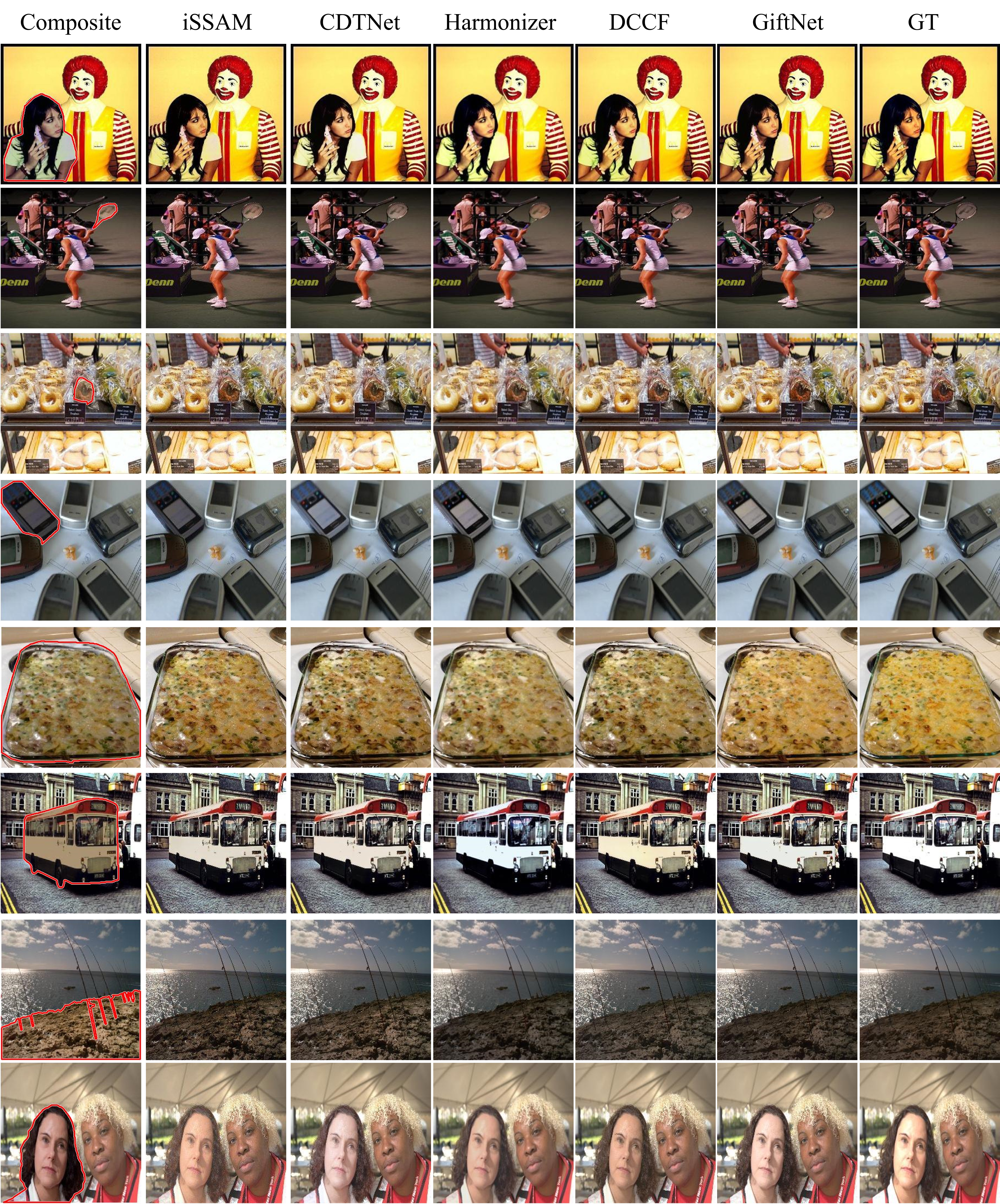}
\end{center}
\caption{From left to right, we show the composite image (foreground outlined in red), the harmonized results of  iSSAM~\cite{sofiiuk2021foreground},  CDTNet~\cite{cong2021high}, Harmonizer~\cite{ke2022harmonizer}, DCCF~\cite{xue2022dccf}, our GiftNet, and the ground-truth on iHarmony4~\cite{CongDoveNet2020} dataset.}
\label{fig:Results_iHarmony4_supp}
\end{figure*}

\begin{figure*}[t]
\begin{center}
\includegraphics[width=0.98\linewidth]{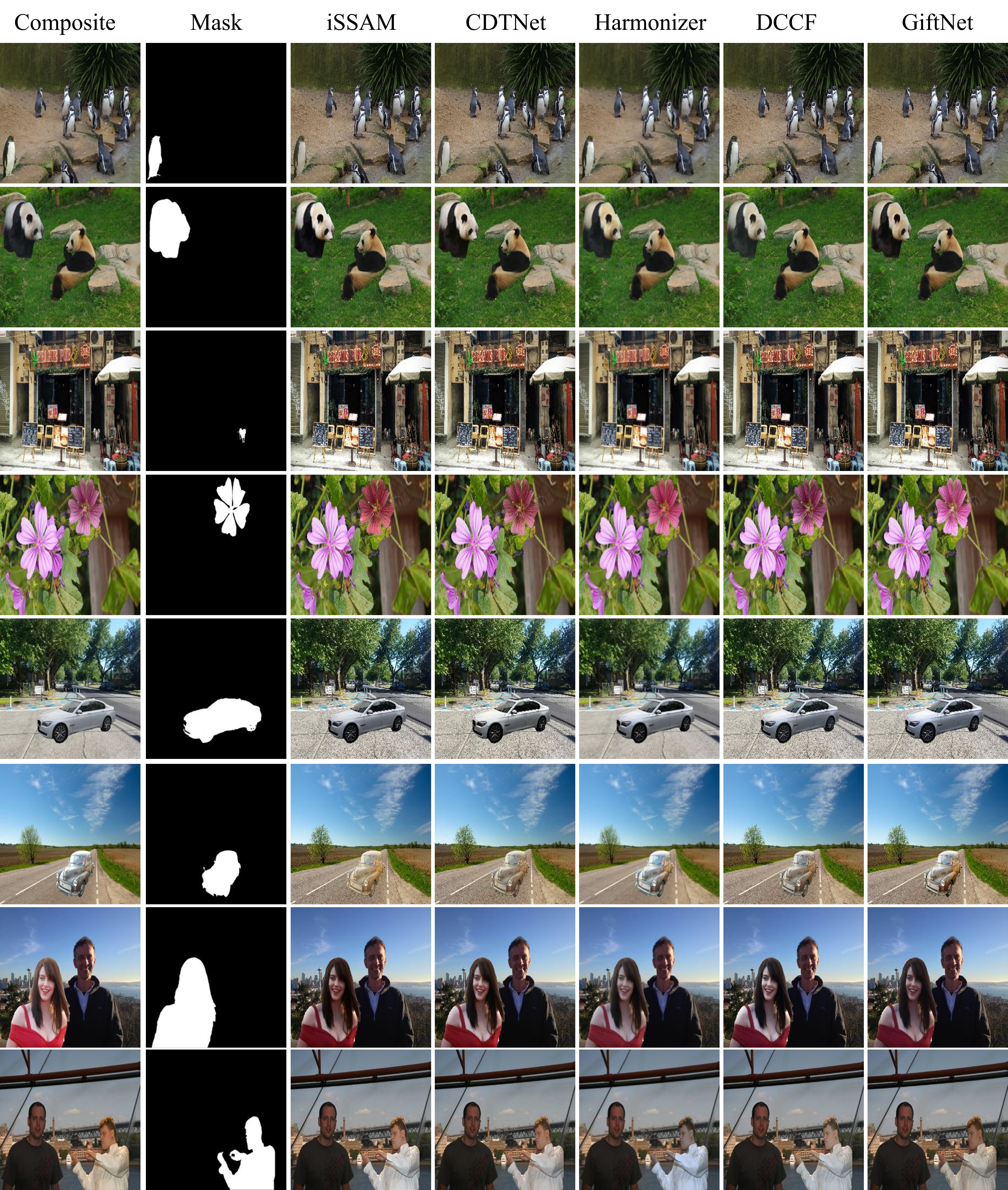}
\end{center}
\caption{From left to right, we show the composite image, the foreground mask, the harmonized results of  iSSAM~\cite{sofiiuk2021foreground},  CDTNet~\cite{cong2021high}, Harmonizer~\cite{ke2022harmonizer}, DCCF~\cite{xue2022dccf}, and our GiftNet on real composite images \cite{cong2021high}.}
\label{fig:Results_HREAL100_supp}
\end{figure*}

\subsection{Foreground Color Transfer}\label{sec:foreground_color_transfer}

As described in Section 4 in the main paper, given an image $\bm{I}_a$, we first  calculate its polynomial matching matrix $\bm{T}_a$ according to its color checker. Then, we apply $\bm{T}_a$ to the foreground $\bm{I}_a^f$ in $\bm{I}_a$ to convert it to $\bm{I}_o^f$, which is expected to be its counterpart in standard illumination condition. In Figure \ref{fig:ccHarmony_standard_illumination}(c), we show several examples of real images $\bm{I}_a$ and their counterparts $\bm{I}_o$  in standard illumination condition.

Next, we randomly select $10$ other real images as reference images for $\bm{I}_a$. For each reference image $\bm{I}_b$, we first calculate its inverse polynomial matching matrix $\bm{T}'_b$ according to its color checker. Then, we apply $\bm{T}'_b$ to $\bm{I}_o^f$ to convert it to $\bm{I}_b^f$, which is expected to be its counterpart in the illumination condition of $\bm{I}_b$.

Finally, we combine $\bm{I}_b^f$ and the background of $\bm{I}_a$ to form a synthetic composite image $\bm{I}_{a\rightarrow b}$. The above procedure has been introduced in the main paper, as illustrated in Figure 2 in the main paper.
Since we select $10$ reference images for each foreground, we can produce $10$ synthetic composite images for each foreground. Based on $426$ foregrounds, we can produce $4260$ pairs of synthetic composite images and real images. 

We split $350$ images into $250$ training images with $308$ foregrounds and $10$ test images with $118$ foregrounds. Thus, the training set contains $3080$ pairs of synthetic composite images and real images, while the test set contains $1180$ pairs. We show several examples of real images and their corresponding synthetic composite images in Figure \ref{fig:ccHarmony_example_composite}.

\section{More Visual Comparison with Baselines} \label{sec:cmp_with_baseline}

We provide more visualization results on iHarmony4 in Figure \ref{fig:Results_iHarmony4_supp}. Compared with the ground-truth real images, our method can usually produce harmonious and visually pleasing results, while the baseline methods may harmonize the composite images insufficiently or incorrectly.

\section{Visual Results on Real Composite Images} \label{sec:real_composite}

The visual comparison between different methods on real composite images is shown in Figure \ref{fig:Results_HREAL100_supp}. We can observe that our method can generally produce more harmonious and realistic images. For example, in row 1-2, the harmonized animals (penguin, panda) of our method are more similar to the same-category animals in the background. In row 3, the harmonized dog of our method is darker since it hides in the darkness. In row 4, the harmonized flower of our method is brighter with vivid color. In row 5-6, the harmonized cars of our method have more appealing and harmonious lustre. In row 7-8, the harmonized portraits of our method are more faithful to the background illumination, by referring to the people standing next to the foregrounds.    

\begin{figure}[t]
\begin{center}
\includegraphics[width=0.99\linewidth]{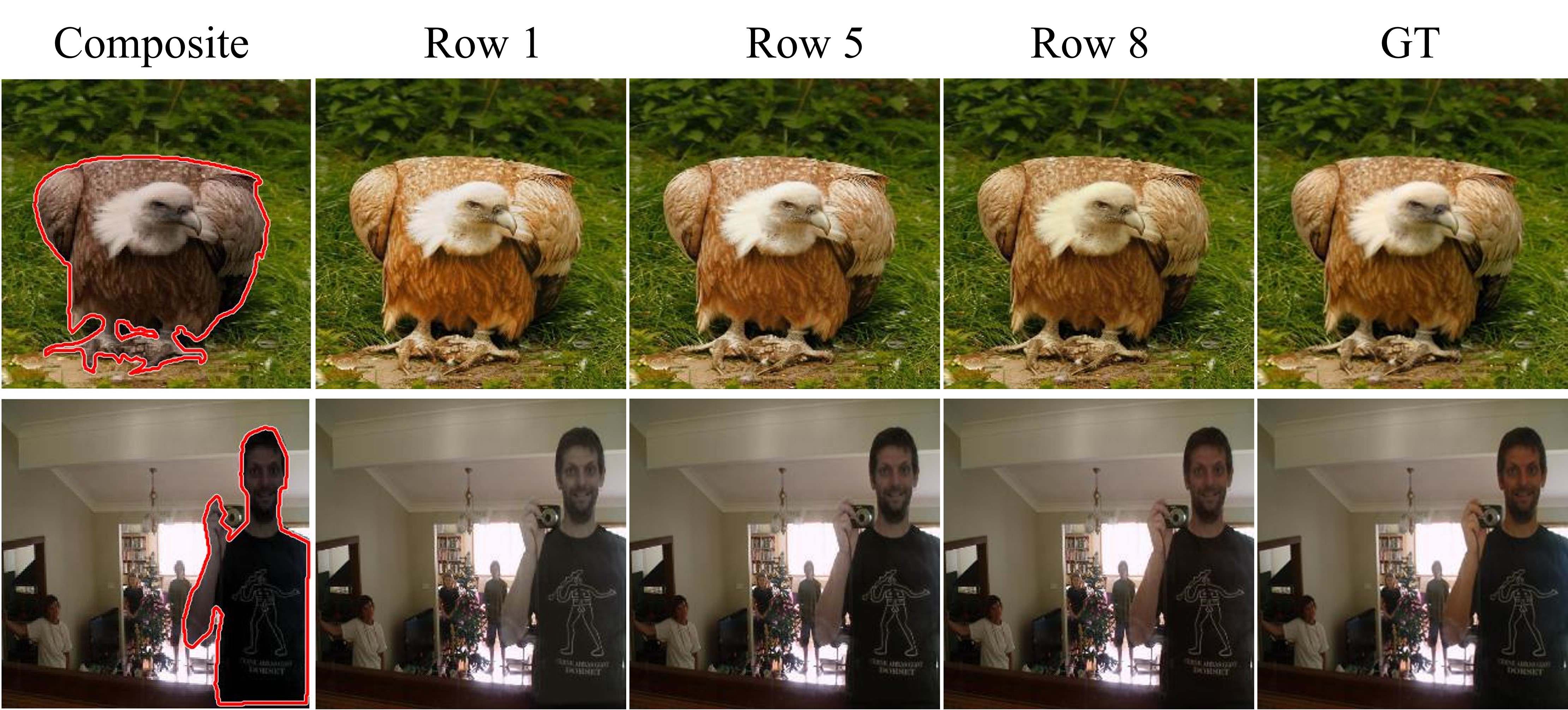}
\end{center}
\caption{From left to right, we show the composite image, the harmonized result of our ablated versions (row 1, 5, 8 in Table 3 in the main paper), and the ground-truth.}
\label{fig:ablation_studies}
\end{figure}

\begin{figure}[t]
\begin{center}
\includegraphics[width=0.9\linewidth]{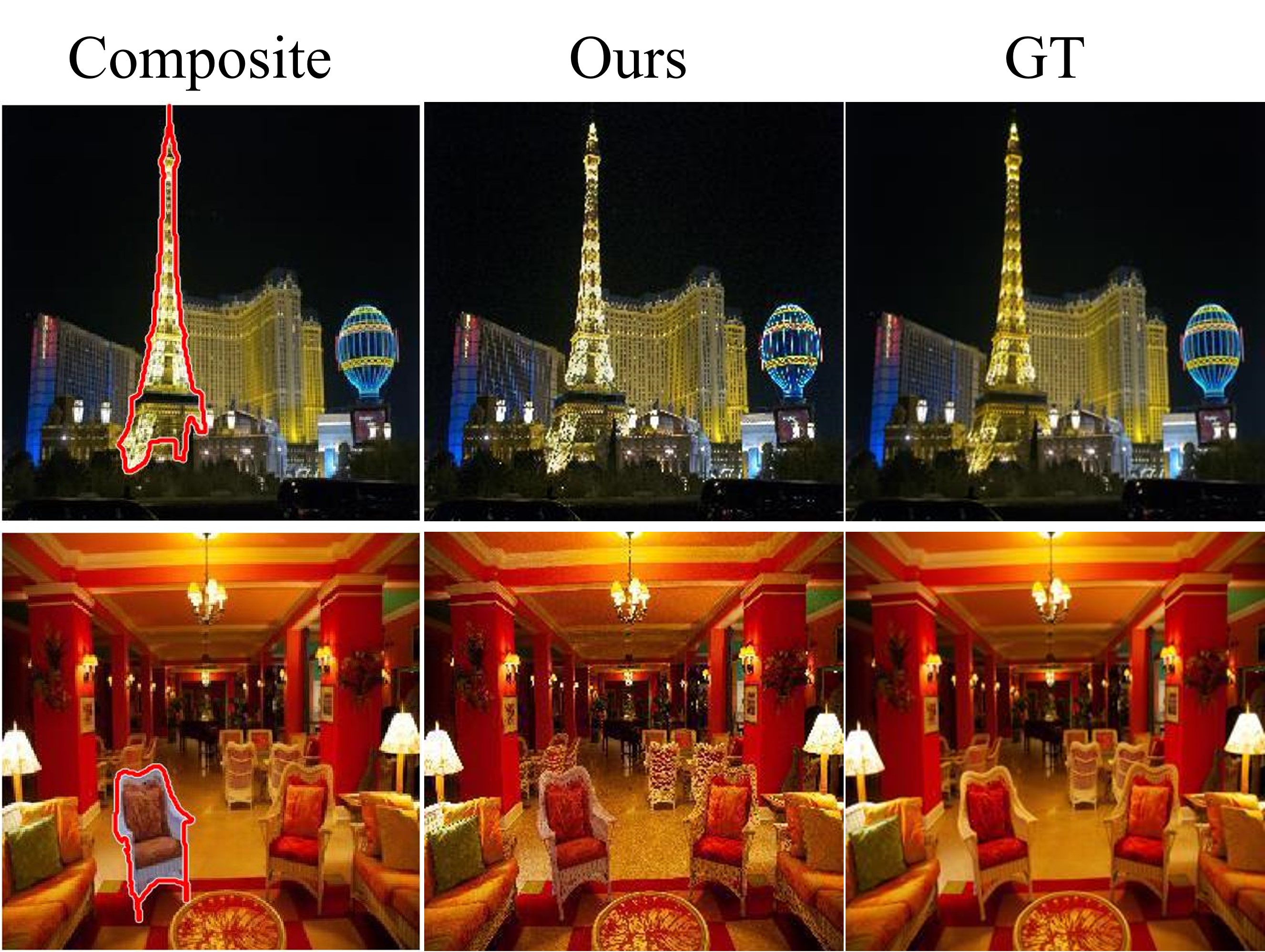}
\end{center}
\caption{From left to right, we show the composite image, the harmonized result of our method, and the ground-truth.}
\label{fig:failure_cases}
\end{figure}

\section{Visualization of Ablation Studies} \label{sec:ablation_studies}

Recall that we have conducted ablation studies for our method in Section 5.3 in the main paper. In Figure \ref{fig:ablation_studies}, we provide the visualization results of several ablated versions of our method, corresponding to row 1, 5, 8 in Table 3 in the main paper. From left to right, we observe that the harmonized images are getting closer to the ground-truth. The results of row 5 are better than those of row 1, demonstrating the effectiveness of our proposed GIFT module. The results of row 8 are better than those of row 5, which proves that relation distillation is useful. 

\section{Failure Cases} \label{sec:failure_cases}
Although our method can usually generate satisfactory harmonized results, there also exist some cases where our method does not behave well. For example, as shown in Figure~\ref{fig:failure_cases}, for the artificial illumination sources, our method fails to produce  satisfactory results, probably because most training images are with natural illumination sources.

\section*{Acknowledgement} The work was supported by the National Natural Science Foundation of China (Grant No. 62076162), the Shanghai Municipal Science and Technology Major/Key Project, China (Grant No. 2021SHZDZX0102, Grant No. 20511100300). 

{\small
\bibliographystyle{ieee_fullname}
\bibliography{camera_supp.bbl}
}